# LOW-RESOURCE LANGUAGE DISCRIMINATION TOWARDS CHINESE DIALECTS WITH TRANSFER LEARNING AND DATA AUGMENTATION

FAN XU, Jiangxi Normal University, China
YANGJIE DAN, Jiangxi Normal University, China
KEYU YAN, Jiangxi Normal University, China
YONG MA, Jiangxi Normal University, China
MINGWEN WANG (CORRESPONDING AUTHOR), Jiangxi Normal University, China

Chinese dialects discrimination is a challenging natural language processing task due to scarce annotation resource. In this article, we develop a novel Chinese dialects discrimination framework with transfer learning and data augmentation (CDDTLDA) in order to overcome the shortage of resources. To be more specific, we first use a relatively larger Chinese dialects corpus to train a source-side automatic speech recognition (ASR) model. Then, we adopt a simple but effective data augmentation method (i.e., speed, pitch, and noise disturbance) to augment the target-side low-resource Chinese dialects, and fine-tune another target ASR model based on the previous source-side ASR model. Meanwhile, the potential common semantic features between source-side and target-side ASR models can be captured by using self-attention mechanism. Finally, we extract the hidden semantic representation in the target ASR model to conduct Chinese dialects discrimination. Our extensive experimental results demonstrate that our model significantly outperforms state-of-the-art methods on two benchmark Chinese dialects corpora.

## 1 INTRODUCTION

Automatic language identification of a speech or a text is the first step of many down streaming natural language processing tasks (i.e., word segmentation, syntactic parsing, discourse analysis, and information retrieval). Although Simoes et al. [29] obtained 97% accuracy on 25 unrelated languages discrimination, the related languages or variations of a specific language is still challenging ([15], [37], [9], [34]).

As is well known, Chinese is spoken in different regions, with distinct differences among these areas. Due to scarce annotation resource, Chinese dialects discrimination is still a challenging task. Recently, in 2018, IFLYTEK organized the first world-level Chinese dialect discrimination contest in order to protect Chinese dialects[1]. They released a 10-way Chinese dialects corpus for the teams participating in the contest. The Chinese dialects corpus consists of 60000 sentences with 35 different speakers covered 10 different Chinese dialects. Similarly, Xu et al. [9] annotated a 19-way Gan (Jiangxi province of China) Chinese dialect corpus. Furthermore, Xu et al. [34] proposed a CNN-based end-to-end language discrimination model toward the Gan Chinese dialects. The total number of sentences, however, is only about 2000. Due to scarce Chinese dialects resource, the language discrimination performance obtains only 60% accuracy on the Gan Chinese dialects corpora.

Since the annotation resource of Chinese dialects is insufficient, there is little specialized research work on Chinese dialects discrimination. Therefore, in this paper, we focus on how to conduct dialect discrimination on the relatively low-resource languages. First, we perform a source-side automatic speech recognition ( ASR ) on a relatively larger Chinese dialects corpus. Then, we augment the target-side low-resource Chinese dialects by using a simple but effective augmentation method (i.e., speed, pitch, and noise disturbance), and fine-tune another target-side ASR model based on the previous source-side ASR model. In addition, we adopt self-attention mechanism to capture the potential common semantic features between source-side and target-side ASR models. Finally, we employ the hidden semantic representation in the target-side ASR model to predict Chinese dialects. Evaluation of two benchmark Chinese dialects corpora shows the appropriateness and effectiveness of the proposed model compared to the state-of-the-art methods.

The major contributions of this paper are three-folds, as follows:

(1) First, we present a novel ASR-based transfer learning framework along with simple but effective data augmentation method to conduct the challenging low-resource Chinese dialects discrimination. We demonstrate the relatively larger source-side ASR model can help to discriminate the fine-tuned target-side Chinese dialects. In fact, these two mechanisms can reinforce one another simultaneously.

(2) Second, the extracted hidden semantic vector matrix, local hidden representation of the input speech feature, in the fine-tuned target-side ASR model can represent the distribution of phonemes corresponding to each frame of a certain speech. Therefore, the phonemes recognized by a certain speech can also help to discriminate the language type of Chinese dialects. Furthermore, in order to extract the potential common semantic features between source-side and target-side ASR models, we successfully extract them by using self-attention mechanism.

(3) Third, we conduct extensive experiments to verify the efficiency of the proposed model on the two benchmark Chinese dialect corpora compared to many strong baseline methods.

The rest of this article is organized as follows. We review related work in Section 2. In Section 3, we present our model CDDTLDA. In Section 4, we present our experiments to investigate the performance of our proposed CDDTLDA, comparing with state-of-the-art approaches, including dataset preparations and experimental result analysis. Finally, we conclude this work and discuss future research in Section 5.

## 2 RELATED WORKS

In this section, we describe representative computational models for similar language discrimination, transfer learning, and data augmentation.

[1] http://challenge.xfyun.cn/2018/aicompetition/tech.

## 2.1 SIMILAR LANGUAGE DISCRIMINATION

Current representative language discrimination models can be classified into two categories: feature-engineeringbased method and neural-network-based approach.

Feature-engineering-based method: More specifically, Tiedemann et al. [31] and Ljubesic et al. [23] adopted a Naive Bayes with unigram features to conduct South Slavic languages discrimination. Moreover, Adorno et al. [10] adopted an SVM and multinomial Naive Bayes with a combination of character N-grams (3≤N≤5) and typed character trigrams to discriminate 14 languages on the DSLCC (Discriminating between Similar Languages Corpus Collection) [2] . What's more, Grefenstette [12] showed bag-of-words features outperformed the syntax or character sequence-based features for English varieties. Also, Elfardyand Diab [7] adopted token level labels and meta features in Arabic varieties discrimination. Similar studies involve Indian languages identification ([22]), Mandarin Chinese and variations identification ([9], [34], [35]), and Persian identification ([21]). Generally, these text-driven feature-engineering-based researches need to design time-consuming features for the language discrimination.

Neural-network-based approach: Due to the success of deep learning, deep learning approaches are adopted in language discrimination. Specifically, Coltekin and Rama [3] presented a deep-neural-network-based model to conduct language discriminating for the similar language DSL (Discriminating between Similar Languages) shared task 2016. Goutte et al. [11] adopted a Convolutional Neural Networks (CNNs) based model with TF-IDF (Term Frequency-Inverse Document Frequency) vectors to conduct similar languages discrimination. Similarly, Xu et al. [34] proposed a CNN-based attention framework for Chinese dialects discrimination. Moreover, Zampieri et al. [36] presented a Long Short-Term Memory (LSTM) based model named deepCybErNet to conduct language discrimination on the DSLCC dataset. Again, Dehak et al. [6] extracted the most salient features in the lowerdimensional i-vector space to conduct language recognition. Furthermore, Snyder et al. [30] proposed a spoken language recognition model based on x-vector. Recently, in 2018, IFLYTEK [13] organized the first worldwide 10-way Chinese dialect discrimination contest. At the same time, they released a LSTM-based baseline system. The champion model in the IFLYTEK 2018 contest consists of the complicated ResNet and Bi-LSTM modules.

## 2.2 TRNASFER LEARNING AND DATA AUGMENTATION

Transfer learning: As a deep learning strategy, transfer learning can reuse knowledge gained from solving one problem by applying it to a different but related problem ([24]). To some extent, it can solve the problem of insufficient data in current popular deep learning-based models. It is widely used in statistical machine learning ([19], [1], [18], [17]). For example, Su et al. [19] proposed a multi-task learning framework to conduct multidomain neural machine translation, using word-level domain context information. In their model, a domain classifier and a adversary domain classifier were presented to classify the input sentences, and an attention-based decoder was adopted to adjust the feedback error in the training process. As to the area of speech analysis, Wu et al. [33] adopted an SVM to train the combination of a range of overlapping character and word n-grams, along with BM25 weight. Then, they conducted the language discrimination on Moldavian as the source language and Romanian as the target language in the transfer task. What's more, Ren et al. [27] focused on low-resource ASR and TTS (Text to speech) tasks simultaneously. They leveraged only few paired speech and text data by integrating denoising auto-encoder, dual transformation, and bidirectional sequence modeling. In addition, transfer learning related to the decoupled speaker characteristics (i.e., speaker embeddings and encoder) can be used in speech synthesis tasks ([16], [32]). Nevertheless, speech-driven ASR-based transfer learning in language discrimination is rare in current literature. Therefore, we are motivated to design ASR-based transfer learning framework to the low-resource Chinese dialects discrimination.

---

[2] http://ttg.uni-saarland.de/resources/DSLCC.

Data augmentation: Data augmentation is another effective way to generate additional training data to meet the requirement of large data in current deep learning-based models. More specifically, Park et al. [26] presented a SpecAugment framework which consists of warping the FBank, masking blocks of frequency channels, and masking blocks of time steps. They successfully applied the SpecAgument for end-to-end speech recognition task. In addition, Chatziagapi et al. [2] adopted the GAN (Generative Adversarial Networks) to generate synthetic spectrograms for overcoming the data imbalance issue, and adopted the generated spectrograms on speech emotion recognition task. Specifically, the generator consists of many transposed convolutional and upsampling layers, while the discriminator includes a list of convolutional layers. Similarly, Etienne et al. [8] demonstrated that the data augmentation with vocal track length perturbation, layer-wise optimizer adjustment can help to conduct speech emotion recognition. Furthermore, data augmentation can also be used in image processing ([28]).

Different from manually designing feature engineering methods and existing deep learning-based approaches, we focus on investigating the scarce Chinese dialects issue, and adopt transfer learning and simple but effective data augmentation which can reduce computational cost to conduct the low-resource Chinese dialects discrimination. In principle, our transfer learning method is a kind of speech-driven method, and the latent hidden representation of speech recognition can help to improve the performance of low-resource language discrimination.

## 3 PROPOSED MODEL

In this section, we explain our Chinese dialects discrimination model in detail.

### 3.1 FRAMEWORK OVERVIEW

Figure 1 shows the framework of our CDDTLDA. As shown in Figure 1, the whole model consists of three modules. The first module is the transfer learning module which is adopted to generate a source-side ASR model by using a relatively larger Chinese dialects corpus. The second one is data augmentation module which is employed to augment the target-side low-resource Chinese dialects. Then, the fine-tuned new target-side ASR model based on the previous source-side ASR model can be used to predict the type of Chinese dialects. Specifically, there are similarities between our target dialect corpus (Gan Chinese dialect and Hakka Chinese dialect corpus) and IFLYTEK corpus. We analyze the geographical location of these two corpora, and find that the geographical range of our target dialect corpus (Gan Chinese dialect and Hakka Chinese dialect corpus) is included in the geographical range of the relatively larger IFLYTEK corpus, so we can learn useful knowledge from IFLYTEK corpus, resulting performance improvement by using transfer learning. Therefore, we use a source-side relatively larger dialect corpus (IFLYTEK) to train a dialect recognition model, then fix the parameters of ResNet, and then adopt our target-side Gan Chinese dialect corpus to fine-tune the model. In fact, the Gan Chinese dialect discrimination can be seen an auxiliary task. The Asr1 in Figure 1 is a model trained with a source-side relatively larger dialect IFLYTEK corpus. Then, the parameters of ResNet in Asr1 are fixed, and the unfixed parameters in Asr1 are fine-tuned with target-side small dialect corpus (Gan Chinese dialects or Hakka corpus) to obtain Asr2. In other words, Asr1 can be taken as a pre-trained model, and Asr2 is a target model obtained by using fine-tuning on the pre-trained model.

In the following paragraphs, we will explain major components inside CDDTLDA, i.e., transfer learning, data augmentation, and classifier.

Transfer learning: Due to the annotation shortage of a specific Chinese dialect, we are motivated to adopt transfer learning to conduct the low-resource language discrimination. As mentioned in the introduction section, IFLYTEK organized the first worldwide Chinese dialect discrimination contest in 2018, and they released a 10- way Chinese dialects corpus which is the only large-scale Chinese dialects annotation resource. Therefore, we employ it to generate a Chinese dialect ASR module. Firstly, we extract speech features (i.e., FBank, MFCC, log-Mels or their combinations) from the initial audio files. Compared with the MFCC ([5]), the FBank has much more information which is only lack of DCT (Discrete Cosine Transform) cepstrum of MFCC. After extracting the speech features, we feed them into our ASR module as shown in Figure 2. For the ASR, we first use a ResNet to extract the local latent representation consists of the abstract speech-level features. Then, a multi-head attention mechanism is performed on the previous latent representation. The function of multi-head attention can focus on the relationship of each frame in a speech to the others. Finally, the output of multi-head attention ( Equation 1) is fed into a fully connected layer with a softmax activation function to generate the final phoneme sequence.

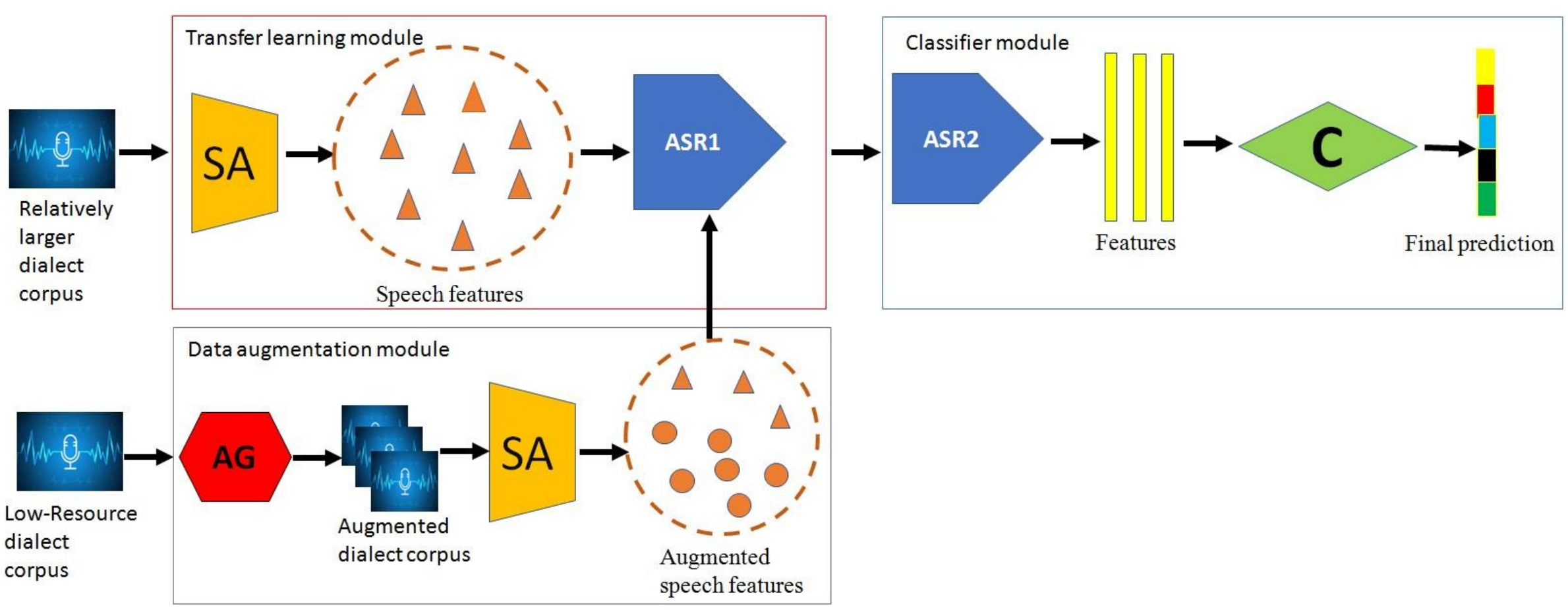

Fig. 1. The overall framework of our CDDTLDA. Here SA stands for signal analysis; ASR refers to automatic speech recognition; ASR1 represents the relatively larger source-side ASR; ASR2 embodies the low-resource target-side ASR; C indicates classifier; AG donates data augmentation.

$$MultiHead(Q,K,V) = Concat(head_1, ..., head_2)W^{o} \quad (1)$$

where

$$head_i = Attention(QW_i^{Q}, KW_i^{K}, VW_i^{V}) \quad (2)$$

and $W_{iQ}$ ⍰ $R(d_{model} \times d_k)$, $W_{iK}$ ⍰ $R(d_{model} \times d_k)$, $W_{iV}$ ⍰ $R(d_{model} \times d_v)$, $W_{iO}$ ⍰ $R(hd_v \times d_{model})$, $h$ refers to the number of heads, and $d_{model}$ is model dimension.

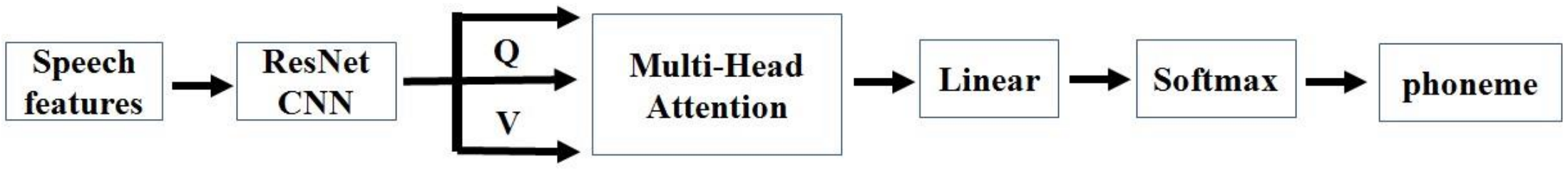

Fig. 2. The framework of ASR

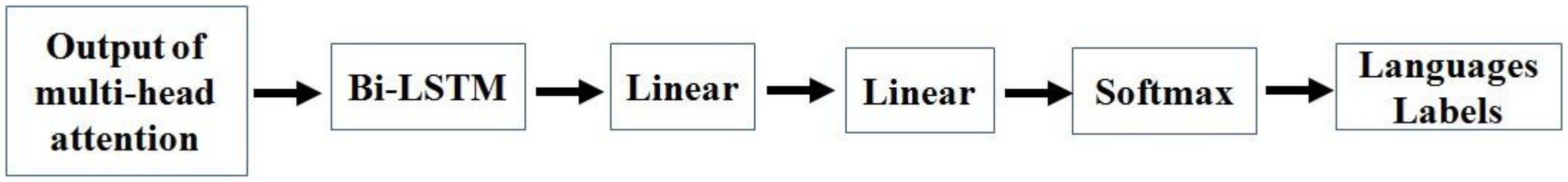

Fig. 3. The framework of classifier

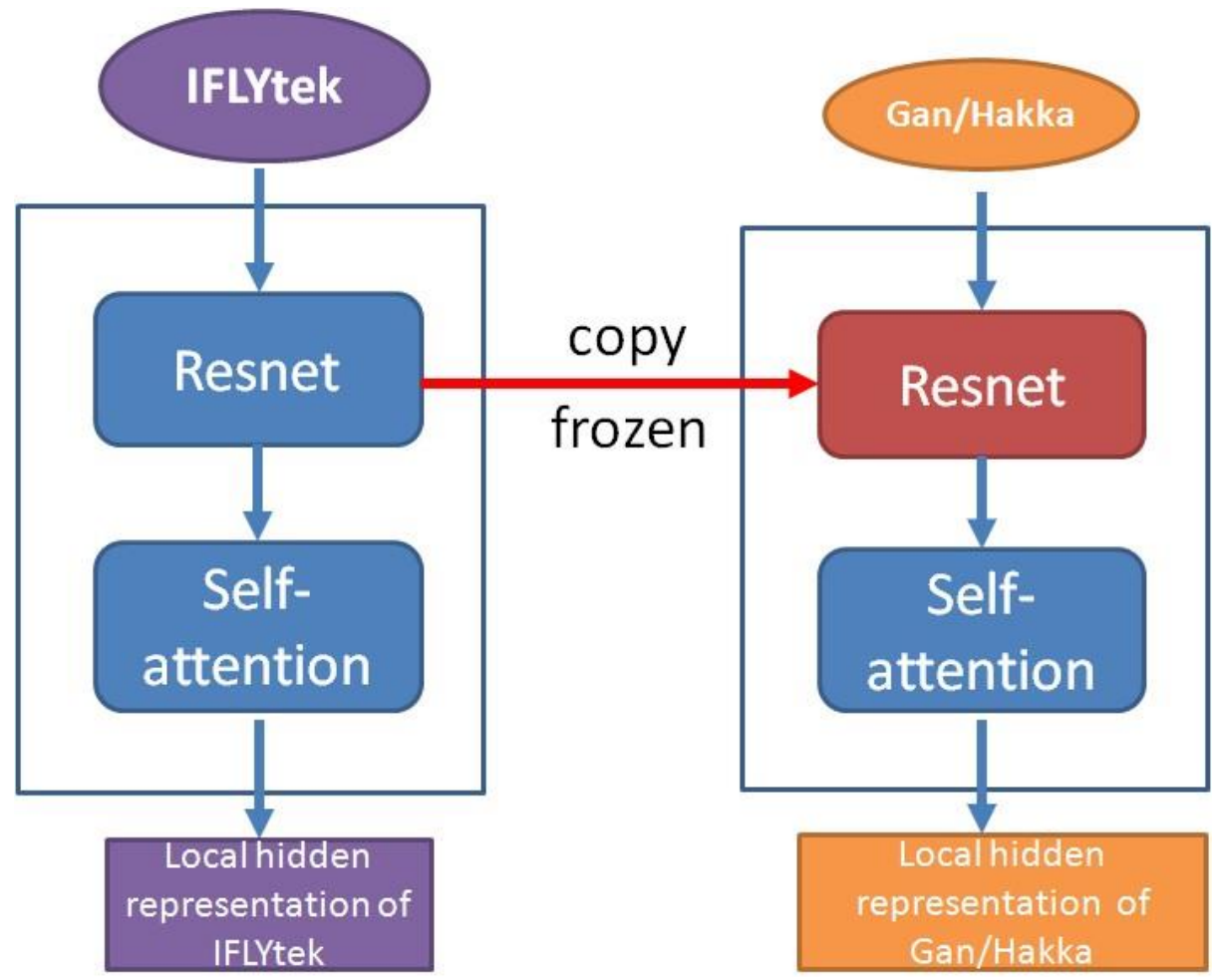

Fig. 4. Execution process of our parameter-based transfer learning.

In addition, we adopt CTC (Connectionist temporal classification) loss as shown in Equation 3 to calculate the difference between the predicted sequence of phoneme and the real phoneme sequence.

$$L_{ctc} = -\ln \prod_{(x,z)\in S} p(z \mid x) = -\sum_{(x,z)\in S} \ln p(z \mid x) \quad (3)$$

where p(z|x) indicates the probability of the output sequence *z* given the input *x*; *S* represents the training set.

Specifically, Figure 4 shows the execution process of our transfer learning method. As shown in Figure 4, we first use IFLYTEK to pre-train our model. Our model parameters include the following parts: ResNet, self attention, and linear. Then, the prediction probability of phoneme sequence is obtained by using softmax activation function. After the pre-training, we obtain the Asr1 model for the relatively larger IFLYTEK dialect corpus. Finally, we fix the parameters of ResNet module in Asr1 model, use our small target-side dialect corpus (i.e., Gan Chinese dialect or Hakka Chinese dialect) to fine-tune the self attention module and linear modules, and obtain the Asr2 model for the target dialect corpus. Since the transfer learning process is related to parameters, we call it parameter-based transfer learning. By contrast, Figure 5 and Figure 6 show instance-based ([4]) and feature representation-based ([25]) transfer learning respectively. As to the instance-based transfer learning, we set same initial weight to each sample for IFLYTEK, and update the weight in each epoch. If a sample does not perform well in the whole training process, we reduce its weight. Otherwise, we will increase its weight. After the training, the weight of suitable sample for transfer learning is high, and the weight of unsuitable sample is low. In the process of training, we take IFLYTEK and our low resource corpus (i.e., Gan and Hakka Chinese dialects corpora) as input to train our ASR model. By contrast, the feature representation-based transfer learning is conducted in a same space. Because the features of IFLYTEK and Gan/Hakka dialect are not distributed in a same space, we design a joint training method to transform them into a same space. We train IFLYTEK speech recognition model and Gan/Hakka speech recognition model at the same time. In the process of training, the parameters of ResNet module in ASR are shared, and a self attention module is connected after the ResNet module. Finally, we add the two losses and update the parameters by a back propagation.

Data augmentation: Although the SpecAugment and GAN-based data augmentation mentioned in related work section has effect on ASR or speech emotion recognition tasks, the potential issue is the high computational cost. In this work, we investigate whether the simple perturbation such as speed, pitch, and noise disturbance can help to discriminate Chinese dialects. At present, we perform three methods for data enhancement, including time extension, pitch conversion and noise addition.

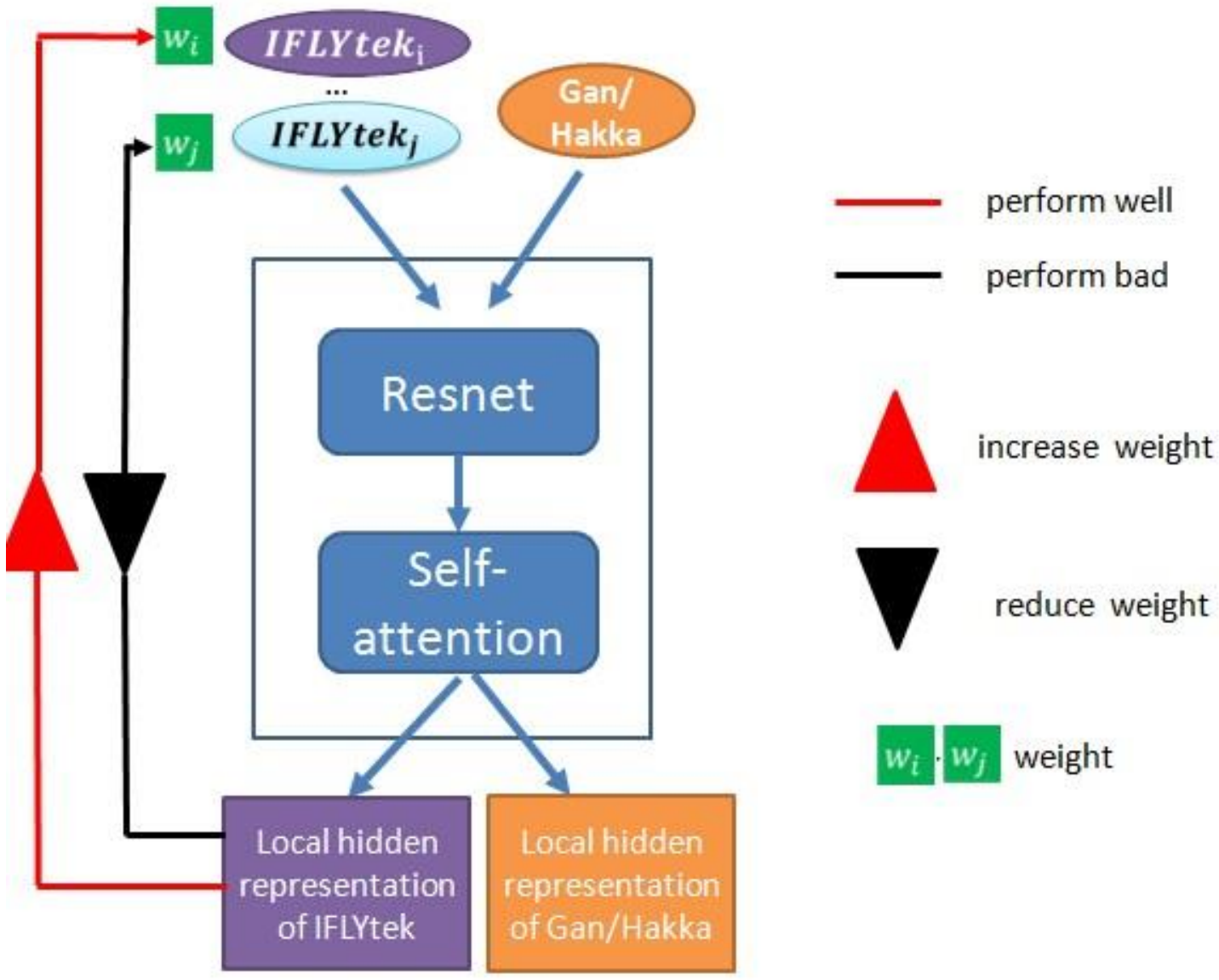

Fig. 5. Execution process of instance-based transfer learning.

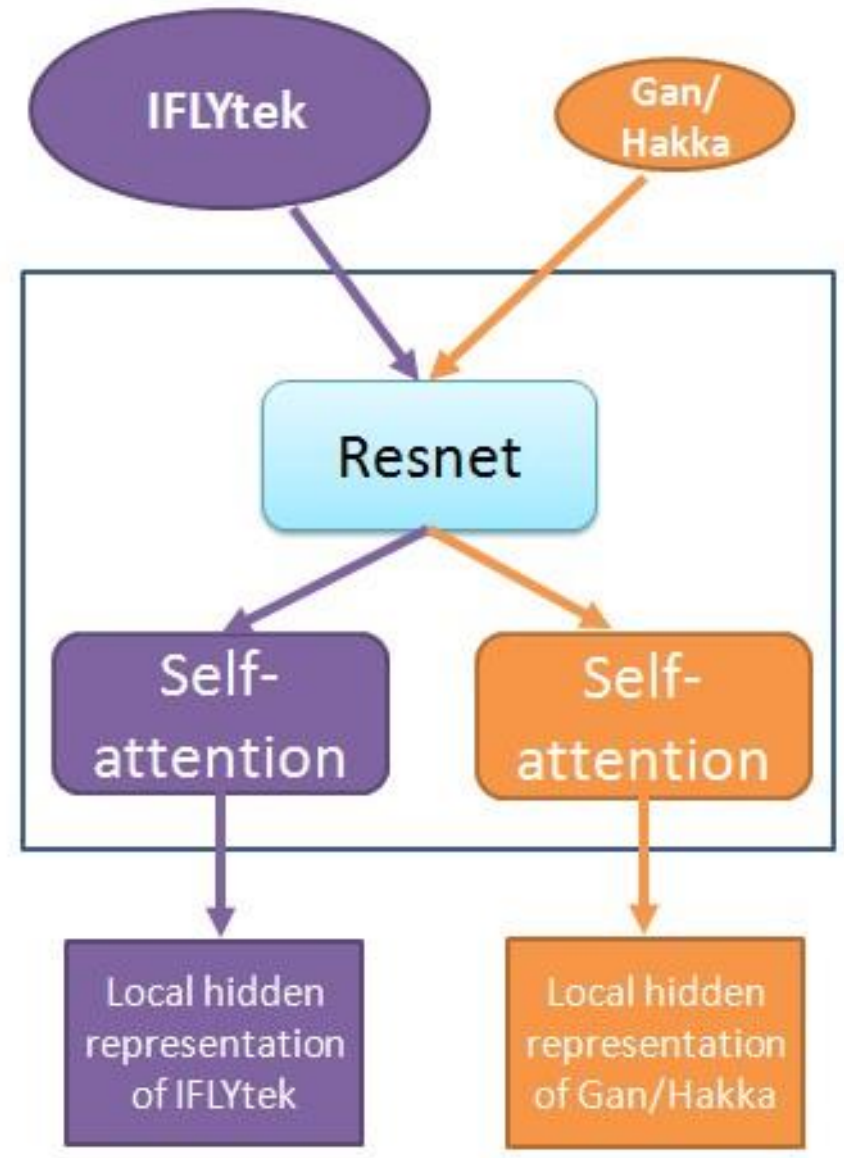

Fig. 6. Execution process of feature representation-based transfer learning.

For the time extension, we slow down or speed up audio sampling while keeping the pitch constant, and stretch the time of each sample by using two factors with {0.9, 1.1}. For the pitch conversion, we increase or decrease the pitch of an audio sample while remaining constant in duration. The pitch of each sample is offseted by eight values (i.e., in semitones) with {−2, −1, 1, 2, −3.5, −2.5, 2.5, 3.5}. For the noise addition, we add random Gaussian noises into previous audio four times. In this case, the mean value of the normal distribution is 0, and the standard deviation of the normal distribution is 1. After using the above three data augmentation methods, the total number of extended audio was 12 times larger than the original sample.

Classifier: As shown in Figure 3, the extracted local hidden representations of the input speech features ( i.e., Fbank, MFCC, log-Mels, or their combinations), which is also an output of multi-head attention in Figure 2 are fed into Bi-LSTM followed by two fully connected layers with a softmax activation function to generate the final type of Chinese dialects. Meanwhile, we adopt cross entropy loss as shown in Equation 4 to calculate the difference between the predicted type of Chinese dialects and the real label.

$$J(\Theta) = \frac{1}{M} \sum_{C \in trainset} -y_c \log[p(y_c = 1)] - (1 - y_c) \log[1 - p(y_c = 1)] + \frac{Q}{2M} \sum_{\vartheta \in \Theta} \vartheta^2 \quad (4)$$

where Θ embodies the parameter sets, *M* donates the number of training instances, and *Q* is adopted to prevent over fitting.

## 4 EXPERIMENTS

In this section, we will investigate the performance of our proposed framework CDDTLDA by comparing with the-state-of-the-art approaches.

### 4.1 DATASETS AND EXPERIMENTAL SETTINGS

In this section, we first briefly introduce the datasets, and then briefly describe evaluation metrics and parameter settings used in our experiments.

DATASETS: As shown in Figure 1, a relatively larger Chinese dialect corpus is employed to generate a source-side ASR model. Therefore, we adopt the IFLYTEK 10-way Chinese dialect corpus as the larger corpus[3] . The corpus is generated within 10 provinces or regions (i.e., ningxia, hefei, Sichuan, shanxi, changsha, hebei, nanchang, shanghai, hakka, and minnan) of China. Each dialect containes an average of six hours of reading-style speech data, covering 35 speakers. Each dialect has 6500 numbers of sentences, totally 65000 sentences. This is the largest Chinese dialects corpus available currently.

In terms of the target-side datasets, we adopt two benchmark corpora. The first one is the 6-way Gan Chinese dialect corpus ([9]) includes "chang jing", "yi ping", "ji lian", "fu guang", "yin yi" and "Hakka". The second one is the 5-way Hakka Chinese dialect corpus includes "Hakka in gan zhou", "Hakka in chang jing", "hai lu", "yue xi", and "yue tai". Some of the Hakka corpus is from GAN dataset, while other Hakka corpus is from the popular dialects website[4]. Table 1 shows the statistics of the two datasets. The former is extracted within only one province of China, and its geographic location is much closer. The latter is generated within different provinces of China, and its geographic location is much looser. Therefore, we would like to verify the influence of geographic factors on Chinese dialect discrimination. We use 80% of them as training set and 20% of them as testing set for each dataset. There are no repeat speakers in training and testing sets.

Table 1. Statistics of the two benchmark datasets.

| Dialect region | Number of Sentences | Total time ( minutes ) |
|---|---|---|
| chang jing | 456 | 77.39 |
| fu guang | 328 | 64.37 |
| gan zhou | 346 | 60.82 |
| ji lian | 477 | 92.66 |
| yin yi | 162 | 36.68 |
| yi ping | 270 | 43.69 |

(a) GAN dataset

[3] http://challenge.xfyun.cn/2018/aicompetition/tech
[4] http://www.phonemica.net/

| Dialect region | Number of Sentences | Total time ( minutes ) |
|---|---|---|
| Hakka in chang jing | 115 | 21.22 |
| Hakka in gan zhou | 212 | 36.85 |
| hai lu | 275 | 23.53 |
| yue tai | 503 | 41.63 |
| yue xi | 67 | 6.98 |

(b) Hakka dataset

Evaluation Metric: We will investigate the performance of our proposed framework and other comparing approaches in terms of four popular evaluation metrics: i.e., accuracy (Equation 5), precision (Equation 6), recall (Equation 7), and F1 (Equation 8). Here, the accuracy is the ratio of the number of sentences with correct type predictions divided by the total number of sentences for the Chinese dialect.

$$Accuracy = \frac{|TP| + |TN|}{|TP| + |TN| + |FP| + |FN|} \quad (5)$$

$$Precision = \frac{|TP|}{|TP| + |FP|} \quad (6)$$

$$Recall = \frac{|TP|}{|TP| + |FN|} \quad (7)$$

where $TP$ indicates true positive, $TN$ stands for true negative, $FP$ donates false positive, and $FN$ refers to false negative.

$$F1 = \frac{2 \cdot Precision \cdot Recall}{Precision + Recall} \quad (8)$$

In addition, we report the system’s performance for Gan Chinese speech recognition using Phoneme Error Rate (PER) which can be computed according to Equation (9) ([20]):

$$PER = \frac{S + D + I}{N} \quad (9)$$

where $S$ indicates the number of substitutions, $D$ represents the number of deletions, $I$ denotes the number of insertions, and $N$ represents the number of phonemes in the corpus.

PARAMETER SETTING: We adopt Kaldi[4] to extract 80-dimension FBank feature, and use librosa to extract 80-dimension log-Mels and 39-dimension MFCC. Table 2 illustrates the parameter setting for ResNet, and Table

[4] http://www.kaldi-asr.org/

Table 2. Parameter setting for ResNet.

| layer | output size | down sample | channels | blocks |
|---|---|---|---|---|
| conv1 | $40 \times L_{in}/2$ | True | 64 | - |
| maxpool | $20 \times L_{in}/4$ | True | 64 | - |
| res1 | $10 \times L_{in}/4$ | False | 64 | 2 |
| res2 | $5 \times L_{in}/4$ | False | 128 | 2 |
| res3 | $3 \times L_{in}/4$ | False | 256 | 1 |
| res4 | $2 \times L_{in}/4$ | False | 512 | 1 |
| mean | $1 \times L_{in}/4$ | False | 512 | - |
| reshape | $512 \times L_{in}/4$ | - | - | - |

Table 3. Parameter setting for self attention layer.

| Model | N | $d_{model}$ | h | $d_k$ | $d_v$ |
|---|---|---|---|---|---|
| Multi-head | 1 | 512 | 8 | 64 | 64 |

3 shows the parameter setting for self attention layer. Furthermore, the value of batch size is set to 10000, the dropout rate is set to 0.1, and the number of epoch is set to 100.

In addition, we adopt Librosa[5] to conduct time extension and pitch conversion, and use the API in Numpy for noise addition. After that, the enhanced voice formats are unified with the audio processing tool SoX[7] by setting sampling rate with 16000 bits. Furthermore, we have implemented the SpecAugment presented by Google, and the two instance-basded and feature representation-based transfer learning methods. In addition, we combined the initial speech features and the generated hidden representation matrix through an addition operation to conduct Chinese dialect discrimination.

## 4.2 BASELINES

To investigate the Chinese dialects discrimination performance of our proposed framework, we will perform comparison studies on following approaches.

LSTM ([13]). IFLYTEK [13] organized the first worldwide Chinese dialect language discrimination contest in 2018. Meanwhile, they presented a strong baseline system based on the LSTM framework. It consists of one hidden layer and takes categorical cross-entropy as loss function with Adam to optimize the loss. What's more, the number of the hidden neuron equals 128, and the learning rate is 0.01.

x-vector ([30]). Snyder et al. [30] proposed a spoken language recognition model based on x-vector. Their model can capture long-term language characteristics in the network by integrating a temporal pooling layer that aggregates information across time. We reproduce their model in this work.

[5] https://librosa.org/doc/latest/index.html
[7]http://sox.sourceforge.net/

ResNet&Bi-LSTM (Champion's model of 2018 IFLYTEK Chinese Dialect Discrimination Contest; [14]). The champion model in the IFLYTEK 2018 contest consists of the complicated ResNet and Bi-LSTM modules. We reproduce their model in our experiment.

CNN_Att ([34]). Xu et al.[34] proposed an attention-driven CNN-based end-to-end model for Chinese dialects discrimination. We reproduce their model in this work.

Table 4. Results of our model using different speech features on the Gan Chinese dialect corpus. Here, T1 indicates our parameter-based transfer learning, A1 stands for our data augmentation method, and hidden donates the local hidden representations matrix.

| | Accu. | Prec. | Rec. | F1 |
|---|---|---|---|---|
| Fbank_T1_A1 | 0.6071 | 0.5248 | 0.6071 | 0.5445 |
| Fbank+hidden_T1_A1 | 0.5548 | 0.4485 | 0.5548 | 0.4678 |
| log-Mels_T1_A1 | 0.5667 | 0.5485 | 0.5667 | 0.4881 |
| log-Mels+hidden_T1_A1 | 0.6095 | 0.6056 | 0.6095 | 0.5403 |
| MFCC_T1_A1 | 0.5929 | 0.5413 | 0.5929 | 0.5273 |
| MFCC+hidden_T1_A1 | 0.5762 | 0.4810 | 0.5762 | 0.5078 |
| Fbank+MFCC_T1_A1 | 0.5429 | 0.4882 | 0.5429 | 0.4907 |
| Fbank+MFCC+hidden_T1_A1 | 0.5762 | 0.6189 | 0.5762 | 0.5338 |
| log-Mels+MFCC_T1_A1 | 0.5810 | 0.4875 | 0.5810 | 0.4994 |
| log-Mels+MFCC+hidden_T1_A1 | 0.5952 | 0.5238 | 0.5952 | 0.5332 |
| Fbank+log-Mels_T1_A1 | 0.5786 | 0.5057 | 0.5786 | 0.5124 |
| Fbank+log-Mels+hidden_T1_A1 | 0.5833 | 0.5833 | 0.5833 | 0.5215 |

## 4.3 RESULTS AND DISCUSSION

In this section, we demonstrate our experimental results on the Gan Chinese dialect corpus (section 4.3.1), results on the Hakka Chinese dialect corpus (section 4.3.2) to investigate: (1) the effectiveness of different speech features (i.e., Fbank, MFCC, log-Mels, or their combinations), and the effectiveness of the combined initial speech features and the output of the extracted local hidden representation. (2) the effectiveness of different data augmentation methods (i.e., our data augmentation like speed, pitch, and noise disturbance; the SpecAgument like frequency and time mask proposed by Google). (3) the effectiveness of our parameter-based transfer leaning compared with other two different transfer learning approaches (i.e., instance-based and feature representation-based transfer learning). In addition, we also report other performance comparison (i.e., effects of number of heads in multi-head attention, parameters size, and PER of Chinese dialects speech recognition) in section 4.3.3, and case study (section 4.3.4).

*4.3.1* RESULTS ON GAN CHINESE DIALECT DATASET. Table 4 lists the results of our model using different speech features on the Gan Chinese dialect corpus. From Table 4, we can obtain following observations:

(1) Compared with the single three speech features (i.e., Fbank, log-Mels, MFCC), the accuracy Fbank achieved is significant higher than other speech features. The log-Mels perform the best in terms of precision on the same corpus.

(2) Compared with the combined speech features, the log-Mels with hidden features perform the best in terms of Accuracy and Recall metrics on the Gan Chinese dialect corpus. The combined initial speech features and the output of the extracted local hidden representation is quite effective to dialect discrimination compared with single speech feature. The combined features can enhance the semantic representation from speech after integrating the output of the multi-head attention in order to capture more local hidden representations in a speech.

Table 5 shows the results of our model using different data augmentation methods with the best speech features on the Gan Chinese dialect corpus. As expected, we can see that the SpecAugment (frequency and time mask) is more suitable for the large Gan Chinese dialect dataset, while our data augment (speed, pitch, and noise disturbance) is effective in the small Hakka Chinese dialect dataset (refer to Table 11). Compared with the complicated SpecAugment, our simple data augment method obtains a comparable performance with the SpecAugment on the larger Gan Chinese dialect dataset. The combination of two data augmentation methods, however, did not obtain better performance due to noises.

Table 5. Results of our model using different data augmentation methods with the best speech features on the Gan Chinese dialect corpus. Here, T1 indicates our parameter-based transfer learning, A1 stands for our data augmentation method, A2 refers to SpecAugment, and hidden donates the local hidden representations matrix.

| | Accu. | Prec. | Rec. | F1 |
|---|---|---|---|---|
| log-Mels_T1_A1 | 0.5667 | 0.5485 | 0.5667 | 0.4881 |
| log-Mels+hidden_T1_A1 | 0.6095 | 0.6056 | 0.6095 | 0.5403 |
| log-Mels_T1_A2 | 0.6071 | 0.5349 | 0.6071 | 0.5428 |
| log-Mels+hidden_T1_A2 | 0.6310 | 0.6769 | 0.6310 | 0.5854 |
| log-Mels_T1_A1+A2 | 0.5857 | 0.5891 | 0.5857 | 0.5639 |
| log-Mels+hidden_T1_A1+A2 | 0.5857 | 0.4495 | 0.5857 | 0.5055 |

Table 6. Results of our model using different transfer learning methods with the best speech features and the best data augmentation method on the Gan Chinese dialect corpus. Here, T1 indicates our parameter-based transfer learning, T2 donates instance-based transfer learning, T3 embodies feature representation-based transfer learning, A2 stands for SpecAugment, and hidden represents the local hidden representations matrix.

| | Accu. | Prec. | Rec. | F1 |
|---|---|---|---|---|
| log-Mels_T1_A2 | 0.6071 | 0.5349 | 0.6071 | 0.5428 |
| log-Mels+hidden_T1_A2 | 0.6310 | 0.6769 | 0.6310 | 0.5854 |
| log-Mels_T2_A2 | 0.5310 | 0.5912 | 0.5310 | 0.5210 |
| log-Mels+hidden_T2_A2 | 0.5881 | 0.5163 | 0.5881 | 0.5388 |
| log-Mels_T3_A2 | 0.5738 | 0.4713 | 0.5738 | 0.5107 |
| log-Mels+hidden_T3_A2 | 0.6024 | 0.4790 | 0.6024 | 0.5278 |

Table 6 illustrates the results of our model using different transfer learning methods with the best speech features and the best data augmentation method on the Gan Chinese dialect corpus. From table 6, we can see that our parameter-based transfer learning approach outperforms the other two instance-based and feature representation-based transfer learning methods. The reason is that our parameter-based transfer learning is a micro method which can capture more detail semantic information in the fine-tuned ASR2 from the pre-trained parameters in ASR1. However, the other two transfer learning (instance-based or feature representation-based ) are more macro instead. Both of them cannot embody much useful information when transferring to dialect discrimination task.

Table 7 shows the performance comparison among baseline systems on the Gan Chinese Dialect Corpus. From Table 7, we can see that our model significantly outperforms all the state-of-the-art approaches on the GAN dataset in terms of all four evaluation metrics. Among the four state-of-the-art methods, x-vector model outperforms all others in terms of all four evaluation metrics. As expected, the x-vector can capture long-term language characteristics by using a temporal pooling layer. The LSTM and CNN_Att models obtain comparable performance. However, the ResNet&Bi-LSTM performs the worst in terms of all four measures. The reason is that the Gan Chinese dialects dataset is much smaller than the IFLYTEK dataset. The complicated ResNet&Bi-LSTM model cannot perform well in this low-resource language discrimination. After performing ablation tests, we know that both transfer learning and data augmentation can help to discriminate Chinese dialects. The hidden semantic distribution of phonemes in the ASR-based model can help to discriminate the type of dialects. Data augmentation can also overcome the scarce annotation issue in some extent.

Table 7. Results on the GAN dataset.

| | Accu. | Prec. | Rec. | F1 |
|---|---|---|---|---|
| LSTM | 0.5310 | 0.3789 | 0.4445 | 0.4038 |
| ResNet&Bi-LSTM | 0.4905 | 0.3438 | 0.4055 | 0.3506 |
| x-vector | 0.5548 | 0.4825 | 0.4720 | 0.4601 |
| CNN_Att | 0.5476 | 0.3997 | 0.4605 | 0.4229 |
| Ours | 0.6310 | 0.6769 | 0.6310 | 0.5854 |
| Ours w/o Transfer learning | 0.5857 | 0.5167 | 0.5857 | 0.5378 |
| Ours w/o Data augmentation | 0.5952 | 0.6223 | 0.5952 | 0.5410 |

Table 8. Performance of Each Gan Dialect Classification.

| | Accu. | Prec. | Rec. | F1 |
|---|---|---|---|---|
| L1 (chang jing) | 0. 7419 | 0. 6832 | 0. 7419 | 0. 7113 |
| L2 (fu guang) | 0. 4118 | 0. 6364 | 0. 4118 | 0. 5000 |
| L3 (gan zhou) | 0. 5455 | 0. 6667 | 0. 5455 | 0. 6000 |
| L4 (ji lian) | 1.0000 | 0. 6443 | 1.0000 | 0. 7837 |
| L5 (yin yi) | 0. 7000 | 0. 4930 | 0. 7000 | 0. 5785 |
| L6 (yi ping) | 0. 0213 | 1.0000 | 0. 0213 | 0. 0417 |

Table 9. The Confusion Matrix of GAN dataset.

| | | Predicted label | | | | | |
|---|---|---|---|---|---|---|---|
| | | L1 | L2 | L3 | L4 | L5 | L6 |
| True label | L1 | 69 | 0 | 0 | 24 | 0 | 0 |
| | L2 | 7 | 28 | 4 | 16 | 13 | 0 |
| | L3 | 4 | 10 | 36 | 6 | 10 | 0 |
| | L4 | 0 | 0 | 0 | 96 | 0 | 0 |
| | L5 | 1 | 6 | 8 | 0 | 35 | 0 |
| | L6 | 20 | 0 | 6 | 7 | 13 | 1 |

More specifically, the performance of our model for each Gan Chinese dialect and confusion matrix is reported in Table 8 and Table 9, respectively. In table 9, L1 stands for “chang jing”, L2 indicates “fu guang”, L3 denotes “gan zhou”, L4 refers to “ji lian”, L5 embodies “yin yi”, and L6 represents “yi ping”. As shown in Tables 8-9, obviously, language discrimination for some Chinese dialects (i.e., “fuguang”,and “yi ping”) , however, is still a challenging task. For example, the dialect “yi ping” is hard to be discriminated since its total number of traing and testing sample is only 223 and 47 respectively.

Table 10. Results of our model using different speech features on the Hakka Chinese dialect corpus. Here, T1 indicates our parameter-based transfer learning, A1 stands for our data augmentation method, and hidden donates the local hidden representations matrix.

| | Accu. | Prec. | Rec. | F1 |
|---|---|---|---|---|
| Fbank_T1_A1 | 0.7393 | 0.7214 | 0.7393 | 0.7094 |
| Fbank+hidden_T1_A1 | 0.6966 | 0.7595 | 0.6966 | 0.6774 |
| log-Mels_T1_A1 | 0.6239 | 0.5177 | 0.6239 | 0.5409 |
| log-Mels+hidden_T1_A1 | 0.6581 | 0.6864 | 0.6581 | 0.6616 |
| MFCC_T1_A1 | 0.7821 | 0.7725 | 0.7821 | 0.7706 |
| MFCC+hidden_T1_A1 | 0.7265 | 0.7629 | 0.7265 | 0.7318 |
| Fbank+MFCC_T1_A1 | 0.7906 | 0.7843 | 0.7906 | 0.7581 |
| Fbank+MFCC+hidden_T1_A1 | 0.7991 | 0.8019 | 0.7991 | 0.7614 |
| log-Mels+MFCC_T1_A1 | 0.7436 | 0.7002 | 0.7436 | 0.7030 |
| log-Mels+MFCC+hidden_T1_A1 | 0.6624 | 0.6686 | 0.6624 | 0.6256 |
| Fbank+log-Mels_T1_A1 | 0.7350 | 0.7137 | 0.7350 | 0.6961 |
| Fbank+log-Mels+hidden_T1_A1 | 0.7607 | 0.7473 | 0.7607 | 0.7443 |

Table 11. Results of our model using different data augmentation methods with the best speech features on the Hakka Chinese dialect corpus. Here, T1 indicates our parameter-based transfer learning, A1 stands for our data augmentation method, A2 refers to SpecAugment, and hidden donates the local hidden representations matrix.

| | Accu. | Prec. | Rec. | F1 |
|---|---|---|---|---|
| Fbank+MFCC_T1_A1 | 0.7906 | 0.7843 | 0.7906 | 0.7581 |
| Fbank+MFCC+hidden_T1_A1 | 0.7991 | 0.8019 | 0.7991 | 0.7614 |
| Fbank+MFCC_T1_A2 | 0.6453 | 0.6355 | 0.6453 | 0.6259 |
| Fbank+MFCC+hidden_T1_A2 | 0.7265 | 0.7340 | 0.7265 | 0.7001 |
| Fbank+MFCC_T1_A1+A2 | 0.6282 | 0.6442 | 0.6282 | 0.6143 |
| Fbank+MFCC+hidden_T1_A1+A2 | 0.6667 | 0.6657 | 0.6667 | 0.6296 |

*4.3.2* RESULTS ON GAN CHINESE DIALECT DATASET. Table 10 lists the results of our model using different speech features on the Hakka Chinese dialect corpus. Compared with the single three speech features (i.e., Fbank, log-Mels, MFCC), the accuracy and F1 MFCC achieved is significant higher than other speech features on the Hakka dialects corpus. Since there are many different accents in each dialect, it is impossible to obtain the best performance using a same underlying speech feature (i.e., log-Mels on Gan Chinese dialects). Each speech feature can help to discriminate different dialects. Again, we also found the combined initial speech features and the output of the extracted local hidden representation outperforms the initial speech features, and the combined Fbank with MFCC obtains the best performance in terms of accuracy, precision, and recall metrics on the Hakka Chinese dialect corpus.

Table 11 shows the results of our model using different data augmentation methods with the best speech features on the Hakka Chinese dialect corpus. Since the size of Hakka dialects corpus is small, our simple data augmentation (speed, pitch, and noise disturbance) outperforms the complicated SpecAugment method.

Table 12 illustrates the results of our model using different transfer learning methods with the best speech features and the best data augmentation method on the Hakka Chinese dialect corpus. Again, our parameter-based transfer learning method outperforms the other two instance-based and feature representation-based transfer learning approaches.

Table 12. Results of our model using different transfer learning methods with the best speech features and the best data augmentation method on the Hakka Chinese dialect corpus. Here, T1 indicates our parameter-based transfer learning, T2 donates instance-based transfer learning, T3 embodies feature representation-based transfer learning, A1 stands for our data augmentation, and hidden represents the local hidden representations matrix.

| | Accu. | Prec. | Rec. | F1 |
|---|---|---|---|---|
| Fbank+MFCC_T1_A1 | 0.7906 | 0.7843 | 0.7906 | 0.7581 |
| Fbank+MFCC+hidden_T1_A1 | 0.7991 | 0.8019 | 0.7991 | 0.7614 |
| Fbank+MFCC_T2_A1 | 0.6838 | 0.7489 | 0.6838 | 0.6617 |
| Fbank+MFCC+hidden_T2_A1 | 0.7009 | 0.7067 | 0.7009 | 0.6647 |
| Fbank+MFCC_T3_A1 | 0.7179 | 0.7283 | 0.7179 | 0.6450 |
| Fbank+MFCC+hidden_T3_A1 | 0.6795 | 0.6536 | 0.6795 | 0.6416 |

Table 13. Results on the Hakka Chinese dataset.

| | Accu. | Prec. | Rec. | F1 |
|---|---|---|---|---|
| LSTM | 0.5812 | 0.4513 | 0.5812 | 0.4787 |
| ResNet&Bi-LSTM | 0.5855 | 0.6893 | 0.5855 | 0.4880 |
| x-vector | 0.6923 | 0.7271 | 0.6923 | 0.6838 |
| CNN_Att | 0.5855 | 0.4895 | 0.6415 | 0.5660 |
| Ours | 0.7991 | 0.8019 | 0.7991 | 0.7614 |
| Ours w/o Transfer learning | 0.5643 | 0.5095 | 0.5643 | 0.5115 |
| Ours w/o Data augmentation | 0.6071 | 0.5428 | 0.6071 | 0.5458 |

Table 13 shows the results on the Hakka Chinese dataset. From Table 13, we can make similar conclusion: our model performs the best in terms of all four evaluation metrics. Again, the ResNet&Bi-LSTM performs the worst. Compared with other baselines, the x-vector model obtains a promising performance. Meanwhile, the ablation tests demonstrate the importance of both transfer learning and data augmentation.

More specifically, the performance of our model for each Hakka Chinese dialect and confusion matrix is reported in Table 14 and Table 15, respectively. In Table 15, L1 stands for "Hakka in chang jing", L2 indicates "Hakka in gan zhou", L3 denotes "hai lu", L4 refers to "yue tai", L5 embodies "yue xi". As shown in Tables 14-15 , obviously, language discrimination for some Chinese dialects (i.e., "yue xi") , however, is still a challenging task. For example, the dialect "yue xi" is difficult to be discriminated since its total number of traing and testing sample is only 51 and 16 respectively.

*4.3.3* OTHER PERFORMANCE COMPARISON. In this section, we give dialect discrimination results varying with different numbers of heads, compare the model size, and lists performance of Chinese dialect speech recognition.

EFFECTS OF NUMBER OF HEADS IN MULTI-HEAD ATTENTION: Table 16 presents the performance with different number of heads on the two datasets (i.e., Gan Chinese dialect and Hakka Chinese dialect corpus). We can know that when the number of heads equals to 8 performs the best in terms of all four evaluation metrics. The more the total number of heads, the more redundant information will exist.

Table 14. Performance of Each Hakka Dialect Classification.

| | Accu. | Prec. | Rec. | F1 |
|---|---|---|---|---|
| L1 (Hakka in changjing) | 0. 9474 | 1.0000 | 0. 9474 | 0. 9730 |
| L2 (Hakka in ganzhou) | 1.0000 | 0. 9254 | 1.0000 | 0. 9612 |
| L3 (hai lu) | 0. 4082 | 1.0000 | 0. 4082 | 0. 5797 |
| L4 (yue tai) | 1.0000 | 0.6273 | 1.0000 | 0. 7709 |
| L5 (yue xi) | 0. 0000 | 0. 0000 | 0. 0000 | 0. 0000 |

Table 15. The Confusion Matrix of Hakka dataset.

|  |  | Predicted label |  |  |  |  |
|---|---|---|---|---|---|---|
|  |  | L1 | L2 | L3 | L4 | L5 |
| True label | L1 | 36 | 2 | 0 | 0 | 0 |
|  | L2 | 0 | 62 | 0 | 0 | 0 |
|  | L3 | 0 | 2 | 20 | 26 | 1 |
|  | L4 | 0 | 0 | 0 | 69 | 0 |
|  | L5 | 0 | 1 | 0 | 15 | 0 |

Table 16. Results with different number of heads.

|  | Accu. | Prec. | Rec. | F1 |
|---|---|---|---|---|
| On GAN dataset |  |  |  |  |
| #heads=4 | 0.5262 | 0.4685 | 0.5262 | 0.4657 |
| #heads=8 | 0.6310 | 0.6769 | 0.6310 | 0.5854 |
| #heads=16 | 0.5667 | 0.5064 | 0.5667 | 0.5111 |
| On Hakka dataset |  |  |  |  |
| #heads=4 | 0.6581 | 0.7120 | 0.6581 | 0.6413 |
| #heads=8 | 0.7991 | 0.8019 | 0.7991 | 0.7614 |
| #heads=16 | 0.6966 | 0.7591 | 0.6966 | 0.6816 |

PARAMETERS SIZE: Table 17 shows parameters size comparison among baseline systems. As shown, although the model size of LSTM and CNN_Att are much lesser than the other three models, the performance of the LSTM and CNN_Att is low. Compared with ResNet&Bi-LSTM and x-vector models, the parameters size of our model is comparable with theirs. However, our model obtains the best performance.

PER OF CHINESE DIALECTS SPEECH RECOGNITION: Table 18 illustrates the PER of Chinese dialect speech recognition for the IFLYtek, Gan Chinese dialect, and Hakka Chinese dialect. From table 18, we can see that the PER on the low-resource Chinese dialects is still quite low and needs further improvement. Generally, Fbank is a quite effective speech feature for Chinese dialects speech recognition. It obtains the best performance on the IFLYtek and Hakka dialects. It obtains comparable performance with the combined Fbank and log-Mels on the Gan Chinese dialects dataset. Although, the performance of Chinese dialects speech recognition is low, the latent representation of ASR can help to detect Chinese dialect discrimination. Because the Chinese dialects speech recognition is not the focus of this work, we will cover the performance improvement as one of our future works.

Table 17. Parameters size comparison.

| Models | Parameters (MB) |
|---|---|
| LSTM | 0.4255 |
| ResNet&Bi-LSTM | 26.6370 |
| x-vector | 17.6132 |
| CNN_Att | 1.2290 |
| Ours on Gan dataset | 52.1167 |
| Ours on Hadda dataset | 56.1216 |

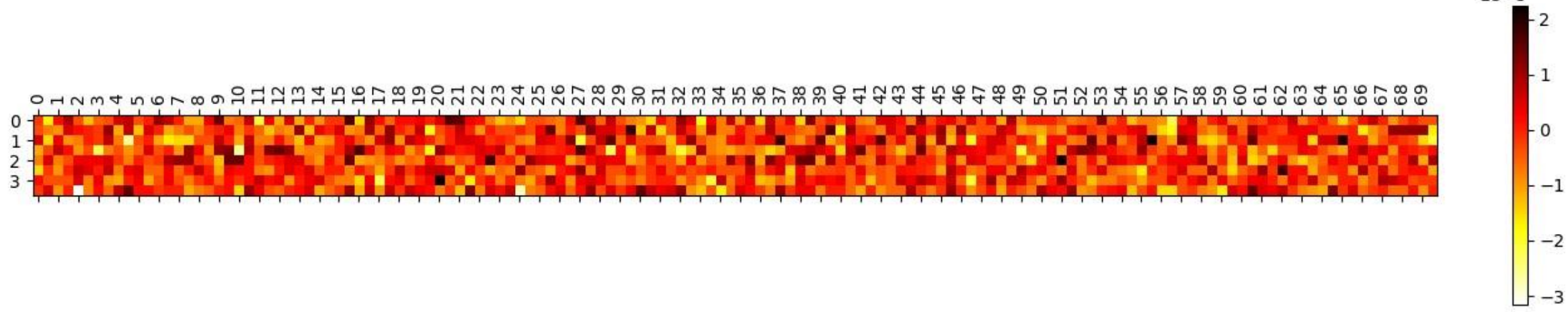

Fig. 7. Attention visualization for an instance from GAN dataset.

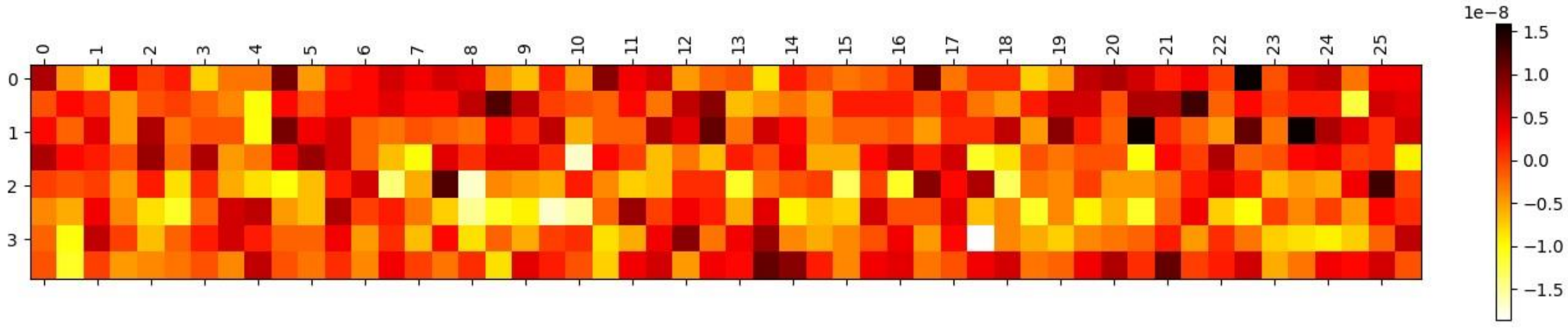

Fig. 8. Attention visualization for an instance from Hakka dataset.

*4.3.4* CASE STUDY. Figures 7-8 show two instances (examples 1-2 represented as phoneme respectively; the number 1, 2, 3 and 4 in the two examples indicate four lexical tones of Chinese) which are only correctly predicted by using our model.

Example 1: b an3 uu uang4 zh e5 b an3 uu uang4 zh e5 d ong1 f eng1 l ai2 l iao3 j vn1 t ian1 d e5 j iao3 b u4 j in4 l iao3

Example 2: g uo2 zh ong1 x i4 zh u2 d ong1 g uo2 zh ong1

As they show, we can observe the attention is useful to conduct Chinese dialects discrimination. From Figure 7 , this audio file is an example in GAN corpus, and the number of frames is 560. After handling by our model, the number of frames is extracted into 140. The x-coordinate 0-69 represents the extracted 140 frames (interval

2) , and the y-coordinate represents the 8 heads of the attention mechanism (interval 2). It can be seen that each head has a certain degree of access to audio information, the first four heads get the most information (with the deepest color), and the discriminative phoneme of the frame number range from 20 to 80 and from 102 to 110 also have a great impact. Similarly, from Figure 8, this audio file is an example in Hakka corpus, the number of frames is 208. After handling by our model, the number of frames is extracted into 52 frames. The x-coordinate 0-25 represents 40 frames after extraction (interval 2), and the y-coordinate represents 8 heads of the attention mechanism (interval 2). It can be seen that the first three heads get the most information (with the deepest color), and the features with frame number ranging from 17 to 28 and from 41-50 also have a great influence. In addition, the discriminated words can be enhanced by using multi-head attention. Since our model adopts transfer learning, it indicates that the features represented in the dark places in Figures 7-8 are some common features of the source-side corpus (IFLYtek) and the target-side corpus (i.e., Gan and Hakka Chinese datasets), resulting the performance improvement accordingly.

Table 18. The Performance of Chinese Dialects Speech Recognition. Here, T1 indicates our parameter-based transfer learning, T2 donates instance-based transfer learning, T3 embodies feature representation-based transfer learning, A1 stands for our data augmentation method, and A2 refers to SpecAugment.

| Models | PER ( % ) |
|---|---|
| on IFLYtek Chinese dialects dataset | |
| Fbank | 0.3912 |
| log-Mels | 0.4158 |
| Fbank+log-Mels | 0.4177 |
| MFCC | 0.4698 |
| Fbank+MFCC | 0.4429 |
| log-Mels+MFCC | 0.4211 |
| on Gan Chinese dialects dataset | |
| Fbank_T1_A1 | 0.5492 |
| log-Mels_T1_A1 | 0.5550 |
| Fbank+log-Mels_T1_A1 | 0.5437 |
| MFCC_T1_A1 | 0.5916 |
| Fbank+MFCC_T1_A1 | 0.5757 |
| log-Mels+MFCC_T1_A1 | 0.5539 |
| log-Mels_T1_A2 | 0.6209 |
| Log-Mels_T1_A1+A2 | 0.5693 |
| log-Mels_T2_A2 | 0.7694 |
| log-Mels_T3_A2 | 0.8069 |
| on Hakka Chinese dialects dataset | |
| Fbank_T1_A1 | 0.7154 |
| log-Mels_T1_A1 | 0.7320 |
| MFCC_T1_A1 | 0.8991 |
| Fbank+MFCC_T1_A1 | 0.7522 |
| Log-Mels+MFCC_T1_A1 | 0.7277 |
| Fbank+log-Mels_T1_A1 | 0.7287 |
| Fbank+MFCC_T1_A2 | 0.8136 |
| Fbank+MFCC_T1_A1+A2 | 0.8597 |
| Fbank+MFCC_T2_A1 | 0.7721 |
| Fbank+MFCC_T3_A1 | 0.8849 |

## 5 CONCLUSIONS AND FUTURE WORK

Low-resource language discrimination is an important task in NLP. In this paper, we present a transfer learningbased framework to discriminate low-resource Chinese dialects along with simple but effective data augmentation. We demonstrate a relatively larger source-side ASR model can help to conduct the fine-tuned target-side lowresource Chinese dialects discrimination. The ASR-based hidden semantic distribution of phonemes corresponding to each frame of a certain speech can help to discriminate the type of dialects. In addition, a self-attention is employed to capture the common semantic features between source-side and target-side ASR models. Our extensive experimental results demonstrate that our model significantly outperforms state-of-the-art methods on two benchmark Chinese dialects corpora.

In our future work, we would like to present other data augmentation methods, and explore more speech-driven features. Furthermore, we will investigate how dialect identification can help other NLP tasks.

## 6 ACKNOWLEDGEMENTS

The authors would like to thank anonymous reviewers for their insightful comments on this paper. This research was supported by the National Natural Science Foundation of China (NSFC) under Grant 61772246, 62162031 , 62066020, and 61876074, Natural Science Foundation of Jiangxi Province under Grant 20192ACBL21030 and 20192AE191005.